 \documentclass[smallabstract,smallcaptions]{dccpaper}
\pdfoutput=1
\usepackage{epsfig}
\usepackage{amssymb,amsmath,amsthm,enumitem}
\usepackage{color}
\usepackage{url}
\usepackage{epsfig}
\usepackage{color}
\usepackage{url}
\usepackage{enumitem}
\usepackage{graphicx}
\usepackage{algorithm}
\usepackage{algorithmicx}
\usepackage{float}
\usepackage[noend]{algpseudocode}
\algnewcommand{\LineComment}[1]{\State \(//\) #1}

\usepackage{titlesec}
\titlespacing*{\section}{0pt}{1\baselineskip}{0.4\baselineskip}
\titlespacing*{\subsection}{0pt}{0.5\baselineskip}{0.2\baselineskip}
\titlespacing*{\subsubsection}{0pt}{0.4\baselineskip}{0.4\baselineskip}
\titlespacing*{\paragraph}{0pt}{0.4\baselineskip}{0.4\baselineskip}

\newlength{\figurewidth}
\newlength{\smallfigurewidth}

\setlist[itemize]{noitemsep, topsep=1pt}
\usepackage[utf8]{inputenc}

\begin{document}

\title
{\large
\textbf{The Sibling Neural Estimator: Improving Iterative Image Decoding with Gradient Communication}
}

\author{Ankur Mali\textsuperscript{$\star$},
Alexander G. Ororbia\textsuperscript{\textdagger},
C. Lee Giles\textsuperscript{$\star$}\\ [0.5em]
\textsuperscript{$\star$\ The Pennsylvania State University, University Park, PA, 16802, USA}\\
\textsuperscript{\textdagger Rochester Institute of Technology, Rochester, NY, 14623, USA}\\
}

\maketitle
\thispagestyle{empty}

\begin{abstract}
For lossy image compression, we develop a neural-based system which learns a nonlinear estimator for decoding from quantized representations. The system links two recurrent networks that ``help'' each other reconstruct same target image patches using complementary portions of spatial context that communicate via gradient signals. This dual agent system builds upon prior work that proposed the iterative refinement algorithm for recurrent neural network (RNN) based decoding which improved image reconstruction compared to standard decoding techniques.
Our approach, which works with any encoder, neural or non-neural,
This system
progressively reduces image patch reconstruction error over a fixed number of steps. Experiment with variants of RNN memory cells, with and without future information, find that our model consistently creates lower distortion images of higher perceptual quality compared to other approaches. 
Specifically, on the Kodak Lossless True Color Image Suite, we observe as much as a $1.64$ decibel (dB) gain over JPEG, a $1.46$ dB gain over JPEG 2000, a $1.34$ dB gain over the GOOG neural baseline, $0.36$ over E2E (a modern competitive neural compression model), and $0.37$ over a single iterative neural decoder.
\end{abstract}

\Section{Introduction}
\label{intro}
Image compression is a fundamental problem that has been at the core of signal and image processing research and applications for decades. Traditional and widely-used approaches such as JPEG and JPEG-2000 rely on hand-crafted codecs and incorporate fixed transform matrices along with quantization and entropy encoders in order to compress images. However, one cannot expect a fixed methodology to obtain optimal solutions for all images with variable content and type.
Recently, deep neural networks (DNNs) have had increasing success in challenging problems such as speech processing,  computer vision, and natural language processing.
Several recent efforts have proposed end-to-end image compression systems using DNNs, achieving impressive results
\cite{balle2016end,rippel2017real,toderici2016full}.
However, while powerful, these systems require carefully designing and training effective encoding/decoding functions as well as quantizers (which are discrete in nature, entailing the further need for the careful design of good smooth approximations of the quantizer \cite{agustsson2017} to facilitate gradient-based optimization). These neural models often rely on large amounts of data and expensive hyper-parameter search, entailing an expensive training process (in terms of computational time and memory).
In contrast, recent work has shown that we may design efficient lossy compression systems by simply combining well-established image encoders with recurrent networks that learn to iteratively estimate a nonlinear decoder, a process known as iterative refinement \cite{orobiadcc}.

In this paper, we build on top of the powerful framework of iterative refinement and propose the sibling neural estimator system. Our approach generalizes the original algorithm to effectively exploit a pair of complementary recurrent neural networks (RNNs) that work together to extract useful information from contextual image patches of different neighboring regions. Importantly, when predicting any given target image patch, these sibling RNNs learn to communicate using their gradient signals through an imbalanced communication channel that encourages one RNN to play the role of source estimator (which extracts rich context information from immediately nearby context patches) while the other plays the role of co-estimator (extracting distant context information that might help the source RNN when reconstructing the target patch, learning to correct its own internal state using the source estimator as a guide). 
Our contributions are as follows:
\begin{itemize}
    \item We propose the sibling neural estimator approach for learned iterative decoding, combining two complementary RNNs via a gradient-based communication channel. Furthermore, we develop a noise-based state regularization scheme and custom optimization scheme to ensure robust training.
    \item We extend and investigate extending our system by developing a scheme where the individual RNNs learn to skip states.
    \item We show that our proposed iterative RNN-based estimator outperforms several baselines as well as the original iterative refinement procedure on a variety of challenging image compression benchmarks.
\end{itemize}

\section{Related Work}
\label{related_work}
Traditional lossy image compression techniques (JPEG, JPEG2000 (JP2)) combine fixed transformation entropy-based encodings with optimized bit allocation in order to achieve better compression at low bit rates \cite{takamura1994coding}. As a result, these approaches are often computationally and memory efficient. Nonetheless, deep neural networks have seen increasing use in developing learnable compression systems. Some research efforts focused on  designing special types of neural networks that focused on predicting images sequentially in two dimensions \cite{oord2016pixel}, which take advantage of bottleneck structure of an autoencoder-like structure, which is well-suited to tackle the problem of compression. As another example, \cite{gregor2016conceptualcompression} proposed a representation learning framework based on a variational autoencoder. Other work achieved state-of-the-art results using spatial-temporal energy compaction \cite{cheng2019learning} or an energy compaction based approach \cite{chen19}, both aiming to extract better latent representations of image data with minimal redundancy in the compresssion model. 

Other proposed methods \cite{Theis2017,rippel2017real,balle2018variational,agustsson2017soft} focus on using a convolution networks or generative adversarial networks (GANs). Recent work on using filter-bank based convolution networks showed improved performance when evaluating on inter-sub band dependencies \cite{dcc19}. 
With respect to utilizing RNNs in general for image compression, the earliest breakthroughs outperformed traditional techniques in terms of visual quality at low bit rates \cite{toderici2015,toderici2016full,johnston2018improved}. These methods proposed end-to-end, differentiable architectures based on convolutional and deconvolutional LSTM hybrid structures. 
Ororbia et.al \cite{orobiadcc} proposed an RNN estimator for nonlinear iterative decoding, combining the custom neural component with traditional encoders, circumventing the need for crafting differentiable approximations of quantization (and was able to take advantage of any efficient encoder design, differentiable or not). More importantly, they showed that such a vastly simple framework could significantly outperform traditional codecs as well as complex end-to-end neural systems.
Our proposed method, the sibling neural estimator, improves upon that work, furthermore reducing the number of iterative refinement steps (yielding better faster inference/test-time decoding in general). Our model, just like theirs, can also be applied to wide range of neural and non-neural image encoders and can handle variable bit-rates.

\section {Nonlinear Estimation through Iterative Decoding and Sibling Models}

\subsection{Iterative Refinement Procedure}

It  reconstructs images from a  compressed  representation and treats it as a multi-step reconstruction problem, which forces the model outside of bad local minima or bad reconstruction over finite samples when trained for K passes.  Patch creation process is adapted from previous work.
Overall the nonlinear estimator is defined by parameters $\Theta = \{\Theta_s, \Theta_t, \Theta_d\}$ and takes in $N$ neighboring patches as input. It is defined by three key components: 
\begin{itemize}[noitemsep] 
\item $\mathbf{e} = e(\mathbf{q}^1, \cdots, \mathbf{q}^N ; \Theta)$, a transformation function of a set of encoded quantized neighboring patches, for a current target patch.
\item $\mathbf{s}_k = s(\mathbf{e}, \mathbf{s}_{k-1} ; \Theta)$, a state function that combines a vector summary of past states with a vector summary of input spatial context.
\item $\widetilde{\mathbf{p}}^j_k = d(\mathbf{s}_k ; \Theta)$, a function that predicts a target over $K$ steps, (or one ``episode''). \footnote{We exclusively use the LSTM for our proposed SNE-RNN system.} 
\end{itemize}
For complete details describing iterative refinement and these functional form refer \cite{orobiadcc}.

\subsection{Functional Forms}
Both the transformation function $\mathbf{e} = e(\mathbf{q}^1, \cdots, \mathbf{q}^N ; \Theta)$ and the reconstruction function $\widetilde{\mathbf{p}}^j_k = d(\mathbf{s}_k ; \Theta)$ can be easily parametrized by multilayer perceptrons (MLPs) as follows. 
\begin{equation}
\mathbf{e} = \phi_e(W_1 \mathbf{q}_1 + \cdots + W_N \mathbf{q}_N) \mbox{, and, } 
\widetilde{\mathbf{p}}_k = \phi_d(U \mathbf{s}_k + \mathbf{c})
\nonumber 
\end{equation}
where $\phi_e(v) = v$ and $\phi_d(v) = v$ (or identity functions). Note that one can significantly cut down number of parameters by tying the block-to-hidden weights, i.e., $W = W_1 = \cdots = W_N$.
For the state function $\mathbf{s}_k = s(\mathbf{e}, \mathbf{s}_{k-1} ; \Theta)$, one may select from a wide variety of recurrent structures, ranging from the classical Elman RNN to the recently popular Long Short Term Memory (LSTM) and other gated RNNs such as the gated recurrent unit (GRU). In preliminary experiments, we used a wide variety of possible memory cell types and determined that LSTM cells yielded best performance overall. Thus, we exclusively use the LSTM for our proposed SNE-RNN system.

\begin{figure}
    \centering
    \includegraphics[width=10.5cm]{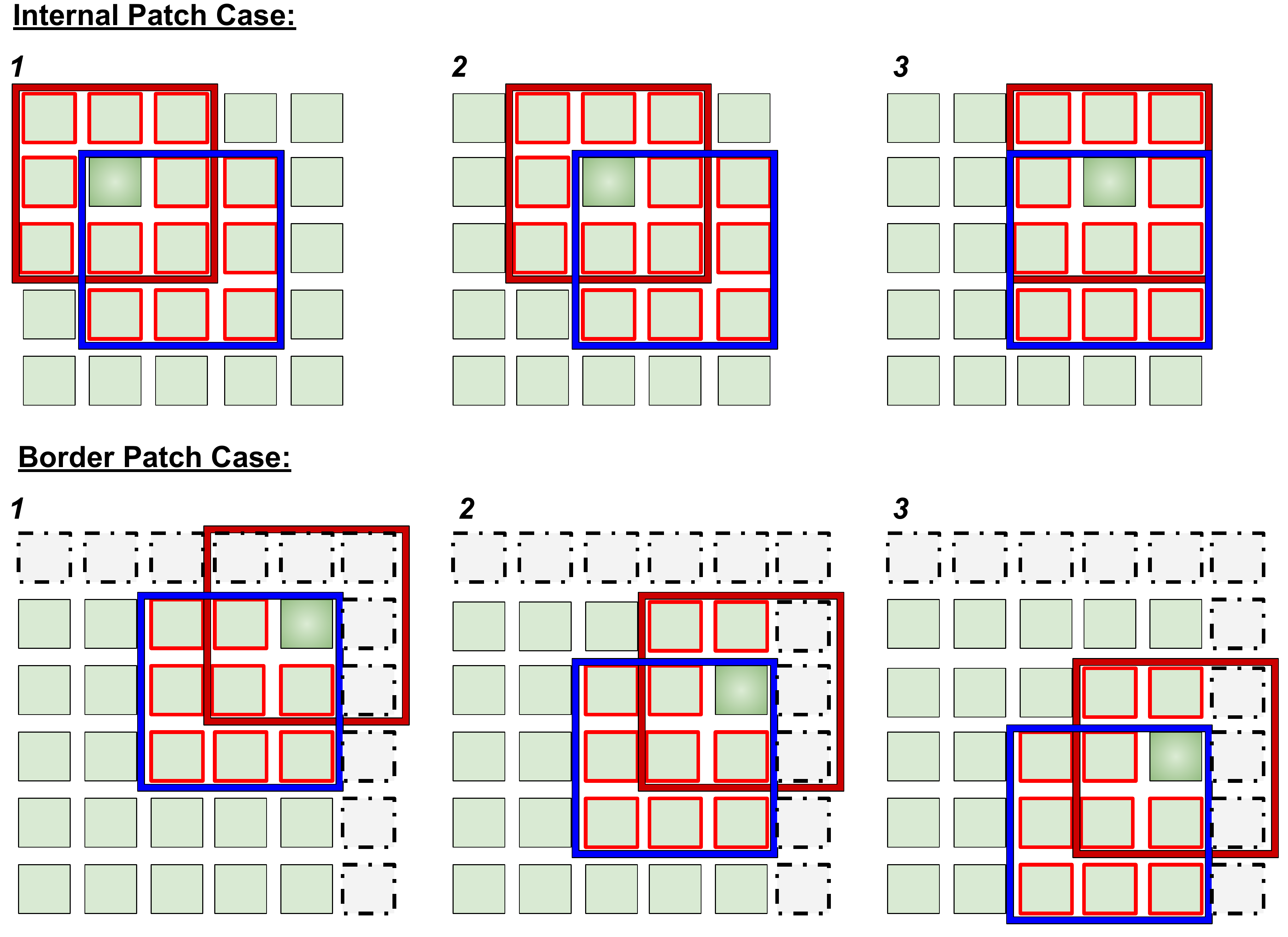}
    \caption{Scan pathway taken by the sibling RNNs. The boxes with dot-dashed boundaries are ``ghost'' patches, or zero-padded image patches. The red box depicts the spatial context selected as input to the central RNN while the blue depicts the spatial context provided as input to the auxiliary RNN. The dark green patch represents the target patch both sibling RNNs aim to reconstruct (which is not fed in as input to either). }
    \label{fig:scan_pathway}
\end{figure}

\subsection{Sibling Neural Decoding System}
\label{twinrnn}

Here we propose a novel generalization sibling neural estimator (SNE).
The intuition is that the original form of iterative refinement only directly fed into limited spatial context as input to the RNN estimator and instead relied on the model's internal state as the primary means to carry information across reconstruction episodes. 
The primary reasons a stateful estimator should be used for image decoding are 
that recurrent weights can be unrolled over a $K$-step reconstruction process and the unfolded network can be viewed as a deep highly nonlinear MLP with weights tied across each step. 
Preliminary experiments showed that a simple extension of the original estimator could force the transformation function $\mathbf{e} = e(\mathbf{q}^1, \cdots, \mathbf{q}^N ; \Theta)$ to take in additional more distant context patches by a small increase in complexity. But this leads to degradation in PSNR by as much as $0.1$ decibels (dB). 
A different approach is that stateful neurons, powerful in helping an estimator learn how to decode, can add an additional stateful estimator to help function approximation. 
Figure \ref{fig:scan_pathway} depicts two typical pathways followed by a pair of sibling neural estimators (the internal patch case and the border patch case, since, according to the original iterative refinement procedure, the estimator predicts border patches last in a circular scan fashion).
Note that our work is different than the bi-directional RNN \cite{graves2005framewise}, since we focused on regularizing the backward RNN and forcing the forward RNN to error correct its signal and improve its long term dependencies.
Formally, at each decoding step, the pair of sibling RNN estimators update their respective hidden states, $z_{t-1}^e$ (for the main estimator) and $z_{t-1}^y$ (for the co-estimator). 
The state update equations for the SNE RNNs, at step $t$, are defined as follows:
\begin{align}
    z_{t}^e =  \o (Ux_{t} + Vz_{t-1}^e)\\
    z_{t}^y =  \o (Ux_{t} + Vz_{t+1}^y)
\end{align}
where $\o$ is non-linear activation function, such as the logistic sigmoid $\o(v) = 1 / (1 + exp(-v))$ or the hyperbolic tangent $\o(v) = (exp(2v) - 1) / (exp(2v) + 1)$ \footnote{note that the above equations assumes Elman-style recurrent units for the sake of illustration, but one could instead use LSTM, GRU, or $\Delta$-RNN units}. Both $z_{t}^e$ and $z_t^y$ contain useful information for predicting the current patch at step $t$, or $x_t$ and the key is to create a communication channel between them. Since we desire a test-time decoding process that is as fast as the original version of iterative refinement, we will have the sibling RNNs exchange gradients through their states. This means that the sibling RNNs will only be linked together at training time, and at test-time decoding, since no gradients are calculated, we will throw away the co-estimator and only use the source estimator, i.e., the co-estimator is only used to improve the reconstruction ability of the source estimator. Furthermore, during training, we want the communication between the two RNNs to be imbalanced -- $z_{t}^e$ should be treated as the richer source of information while $z_{t}^y$ is treated as only a helpful, auxiliary information source meant to enrich $z_{t}^e$. Having specified the two qualities we want for the communication between the sibling estimators, we can now define how gradients are passed:
\begin{align}
    H_{Comm} =   || (W_{err} z_t^y) - z_t^e || \mbox{.}
\end{align}
Note that the above is simply a weighted euclidean distance function that measures how far off the co-estimator $z_t^y$ is from the source estimator $z_t^e$. Since the communication $H_{Comm}$ is meant to be imbalanced, we have introduced an additional parameter matrix $W_{err}$, i.e., the error correction matrix which functions as a linear correction factor for the co-estimator RNN. Furthermore, $W_{err}$ serves to ``weaken'' the co-estimator state $z_t^y$ which prevents the degenerate case of both RNNs collapsing to identical state values (which would lead to unstable training given that both estimators take in as input different context patches). Crucially, observe that the above distance function works in two directions -- during training, the gradients calculated for the source estimator will traverse through this channel into the co-estimator and vice versa.

\begin{figure}
    \centering
    \includegraphics[width=14.00cm]{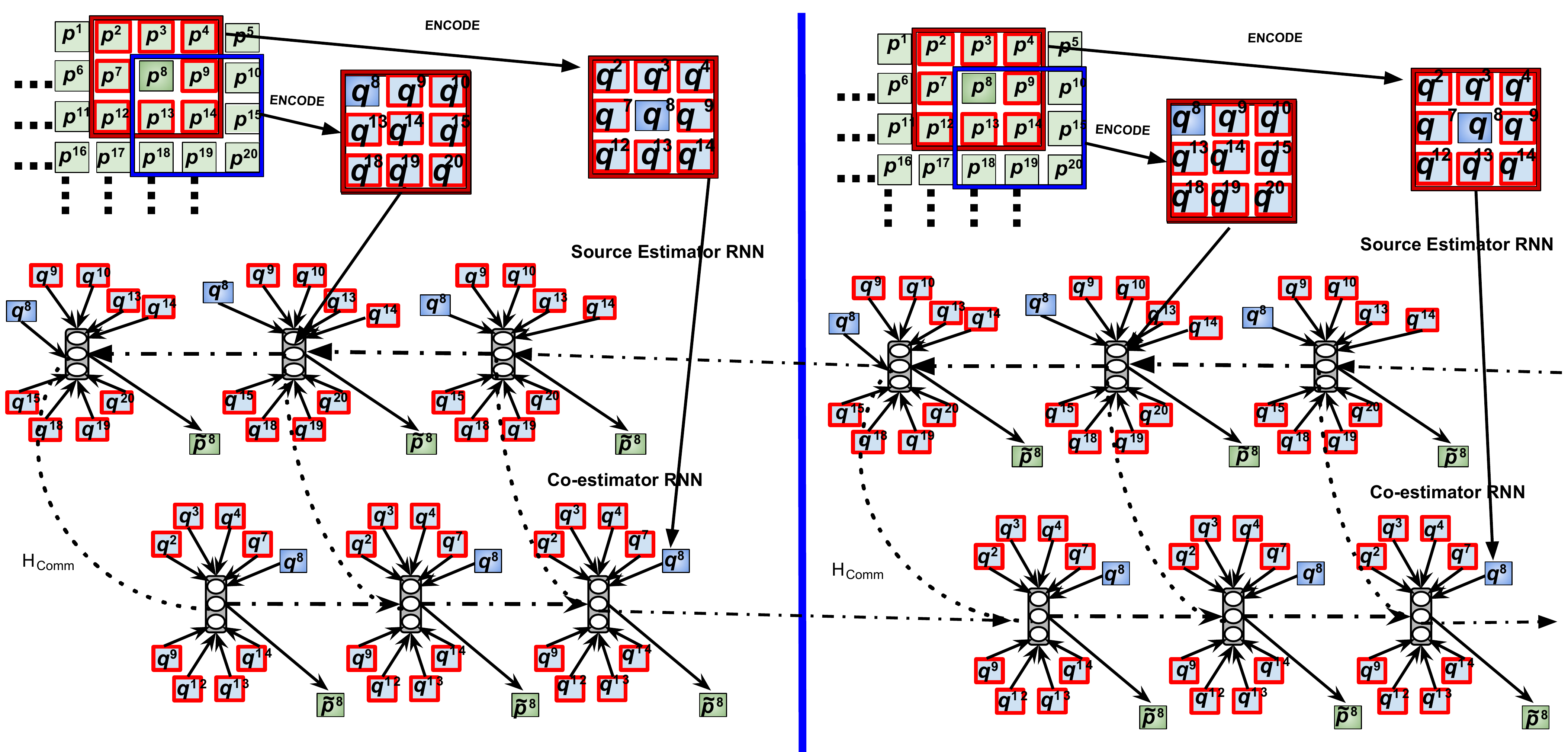}
    \caption{Iterative refinement ($2$ episodes, $K=3$) for the sibling RNN estimator system (top-level view). Dot-dashed arrows indicate the latent state across an episode. $H_{Comm}$ is the communication link between the source estimator and co-estimator RNNs. The dashed arrow indicates that the reconstruction memory carried across each step within an episode. The source estimator RNN takes in as input the set of patches in the red boxed zone while the co-estimator RNN takes in as input the set of patches in the blue boxed zone.}
    \label{fig:Model_diagram}
\end{figure}
To furthermore improve the generalization ability of our proposed sibling RNN system, we also designed a noise-based regularization scheme where an adaptive variance parameter is used to drive the amount of noise based on the epoch index (such an adaptive scheme has proven to be far more effective than simple gradient noise or L2/L1 decay when regularizing more complex, higher-order RNNs \cite{mali2019neural}). Under this scheme, we pick an initial value for the  variance at the first epoch and lower it as a function of epoch. While this scheme has been applied directly to the synaptic weights in prior work, interestingly enough, we found that directly regularizing the state of (only) the source estimator RNN proved to be far more effective. We can rewrite $H_{Comm}$ as follows:
\begin{align}
    H_{Reg-Comm} = ||( I(z_t^f) - (z_t^b + \mathcal{N}(\mu, \sigma^2) ))
\end{align}
where $\mu$ (mean) and $\sigma^2$ (variance) are user-set meta-parameters to control the strength of the state regularization noise. Our regularization schedule proceeds as follows: for every $8$th epoch, we apply use the regularized channel $H_{Reg-Comm}$ instead of $H_{Comm}$ to link the sibling RNNs (and set the number of iterative refinement steps to $K=3$, while for the other epochs, we use $H_{Comm}$ instead (with number of refinement steps set to $K=2$).
After $120$ epochs, we stop using the adaptive noise state regularization scheme. 
To optimize our model parameters, we define our loss function to be the mean square error (MSE) to measure the mismatch between each estimator's reconstructed patch and the target patch $\mathbf{p}^j$. Formally, the reconstruction loss (over a $K$-step iterative refinement episode) is defined (over a mini-batch of $B$ patches) as follows:
\begin{align}
    \mathcal{D}_{MSE}(\mathbf{p}^j,\widehat{\mathbf{p}}^j) = \frac{1}{(2 B K)}  \sum^K_{k=1} \sum^B_{b=1} \sum_i ( \widetilde{\mathbf{p}}^{j,b}_k[i] - \mathbf{p}^{j,b}[i] )^2 \mbox{.}
\end{align}
We then apply this reconstruction loss to both the source and co-estimator RNNs. The total objective function then consists of the two reconstruction loss terms as well as the communication channel term (in this case $H_{Comm}$ but, for certain epochs, as described for our noise scheme, we temporarily swap in $H_{Reg-Comm}$), defined as:
\begin{align}
    L= \mathcal{D}_{MSE}(\mathbf{p}^j,\widehat{\mathbf{p}}^{j,y}) + \mathcal{D}_{MSE}(\mathbf{p}^j,\widehat{\mathbf{p}}^{j,e}) + \alpha H_{Comm}
    \label{cost_fun}
\end{align}
where $\alpha$ is an externally set coefficient to control the strength of the communication channel. We label the prediction of target patch $\mathbf{p}^j$ differently for each sibling estimator, i.e., $\mathbf{p}^{j,y}$ is the co-estimator's prediction while $\mathbf{p}^{j,e}$ is the source estimator's prediction (note that we do not include the $\mathcal{D}_{MAE}$ of the original algorithm in \cite{orobiadcc} as we found this actually worsened performance for a the sibling estimator system). Backpropagation through time is used to calculate gradients and, as a result of extensive preliminary experimentation, we found that a stable optimization process for updating the sibling estimator weights should proceed as follows:  for $120$ epochs we update weights using the Adam update rule and then switch (at the moment we shut off our state regularizer) to stochastic gradient descent to conduct a gentle fine-tuning phase.
Again, we reiterate that at test time decoding, we simply run the iterative refinement procedure of \cite{orobiadcc} using only the source estimator RNN (giving us the same test-time decoding computational complexity as in \cite{orobiadcc}) , i.e., after training, we discard the co-estimator RNN and only decode image patches using the source estimator RNN. 

\subsection{Learning to Skip the States}
\label{sec:skip-states}
Finally, we experimented with a variation of our RNN estimators based on the ``SkipRNN'' proposed in \cite{skiprnn}. SkipRNN, in short, entails incorporating a mechanism into the RNN's computation that essentially allows the model to ``skip'' states as it processes a sequence. This is done to ultimately shorten the length of the underlying computational graph when having to perform back-propagation through time. Since our iterative decoding system employs two RNNs during training, utilizing such a mechanism should help us to reduce the computational cost without compromising model generalization ability.
We experiment with three ways of utilizing the SkipRNN approach. First is SNE-RNN-SkipB, in which the system must only learn to skip the states of the source estimator RNN. The second variation is SNE-RNN-SkipF, which entails having the system to learn to only skip the states of the co-estimator RNN. 
Finally, we propose SNE-RNN-SkipBoth, which means the system learns to skip states for both of sibling RNNs. 
In our implementation of the SkipRNN, we augment a network with a binary state update gate, $u_t \in \{0,1\}$, where $0$ means we copy (or ``skip'') the previous time step and $1$ represents the choice to update the current state. The SNE-RNN-SkipBoth's update equation is defined as:
\begin{align}
    u_t^f &= f_{bin}(\hat{u_t^f}), \quad  u_t^b = f_{bin}(\hat{u_t^b}) \\
    z_t^f &=u_t^f.\o(Ux_{t} + Vz_{t-1}^f) + (1-u_t^f). z_{t-1}^f \\
    z_t^b &=u_t^b.\o(Ux_{t} + Vz_{t-1}^b) + (1-u_t^b). z_{t-1}^b \\
    \triangle \hat{u_t^f} &= \sigma(W_p^f z_t^f + b_p^f), \quad  \triangle \hat{u_t^b} = \sigma(W_p^b z_t^b + b_p^b) \\
    \hat{u_{t+1}^f} &= u_t^f. \triangle \hat{u_t^f} + (1-u_t^f).(\hat{u_t^f} + min(\triangle \hat{u_t^f},1-\hat{u_t^f})) \\
    \hat{u_{t+1}^b} &= u_t^b. \triangle \hat{u_t^b} + (1-u_t^b).(\hat{u_t^b} + min(\triangle \hat{u_t^b},1-\hat{u_t^b}))
\end{align}
where $W_p^f$ and $W_p^b$ are weight vectors for co-estimator RNNs and source estimator RNNs, respectively. $b_p^f$ and $b_p^b$ represent biases and $\sigma(v) = 1 / 1 + exp(-v)$. $f_{bin}$ is the binarizer (0 or 1). For all operations, except $f_{bin}$, are end to end differentiable. For the $f_{bin}$ function, we use the straight-through estimator to approximate gradients during back-propagation.

\section{Experiments}
\label{experiments}
In our experiments, we implement the SNE variations presented above and compare them to JPEG and JPEG 2000 (JP2) baselines as well as an MLP stateless decoder and the original iterative refinement procedure of \cite{orobiadcc}.  In addition, we compare to the competitive neural architecture proposed in \cite{toderici2016full} (GOOG) as well as the state-of-the-art variational compressor E2E \cite{balle2016end} on all of our test-sets. 
We present compression results for all models on six challenging image compression benchmark used in \cite{orobiadcc}, i.e., Kodak, CB~8-Bit, CB~16-Bit, CB~16-Bit-Linear, Tecnick, and Wikipedia. For a complete description of the compression benchmarks, we refer the reader to \cite{orobiadcc}.

\subsection{Data and Benchmarks}  
\label{exp:data}
The training set contained randomly sampled high resolution 128k images from the \emph{Places365} {\cite{zhou2017places}} dataset, down-sampled to $512\times512$ pixel sizes. We randomly sampled 450k images from Imagenet and resized them into $224\times224$. In addition, we also randomly sampled 7168 raw images with variable bit rates from the RAISE-ALL{\cite{Raise}} dataset, which were then down-sampled to $1600\times1600$ dimensions. The first step is to compress each image with variable bit rates in range $0.35$--$1.02$ for any given encoder (once for JPEG and once for JP2). Similarly, to create a validation sample, we randomly selected 20K images from the \emph{Places365}, 35k from imagenet and 1k from RAISE-ALL combined together to create a single set. Validation bit rate for each image is constant with the training data. We then divide images into set of non-overlapping patches (JPEG) and set of overlapping patches (JP2). 
We experimented with $6$ different tests based on prior work \cite{orobiadcc}.

\subsection{Experimental Setup}
\label{exp:design}
All of our SNE-RNN models contained one layer of $512$ hidden units. We randomly initialized weights from a uniform distribution,$\sim U(-0.05, 0.05)$. Parameters were updated during training using the optimizer switching heuristic we described earlier, estimating gradients over mini-batches of $512$ samples and hard clipping to a magnitude of $15$ (to prevent exploding gradients). We started training with Adam with an initial learning rate of $2$e-$4$ and used polynomial decay until the $120$th epoch. 
After the $120$th epoch, we switch to stochastic gradient descent (SGD) and continue our training up to $300$ total epochs. This approach, as we described earlier, exploits the faster convergence properties of Adam with the fine-tuning capabilities of SGD (note that this approach also performs better than stochastic annealing learning rate used in prior work on neural-based decoder estimation). We also use the two-step shuffling approach outlined \cite{orobiadcc} to ensure robust training.

\subsection{Results}
\label{sec:results}
We present our model compression results for the $6$ benchmark datasets mentioned above in Table \ref{results:benchmarks}.
Each model was evaluated using three popular metrics \cite{ma2016group}: PSNR, structural similarity (SSIM), and multi-scale structural similarity (MS-SSIM \cite{wang2004imagequality}, or $MS^3IM$ as shown in our tables). 
In table \ref{results:benchmarks} we observe that our proposed SNE approach yields the best results. Our approach consistently yields better performance when compared with classical models (JPEG, JP2), competing end-to-end neural compression models (GOOG, E2E) and even the original iterative refinement procedure (LSTM-JPEG , LSTM-JP2). We tested on 6 different independent benchmarks, which contain unseen and out of sample images to test the true generalization performance. We report the scores for 3 popular metrices PSNR, SSIM and MS-SSIM, and show that our decoder performance is consistent across all metrics.
In Table \ref{results:kodak_progressive}, we show how PSNR varies as the number of reconstruction steps $K$ allowed per episode varies. To create this table, we randomly sample 4 images from our Kodak test suite and report the PSNR measurement for our best SNE model as a function $K = \{1,2,3,7,9,11\}$. We observe that for our SNE model, $K=2$ or $K=3$ is sufficient to achieve better performance. This is much better when compared to earlier work \cite{orobiadcc}.
Note that our results are the best or comparable to GOOG and E2E, and our system is much simpler in terms of model complexity, i.e., number of parameters, and requires dramatically less training time.

\section{Conclusions}
We proposed a new neural-based system for lossy image compression, the sibling neural estimator (SNE) system. Variants of our proposed model under the \emph{iterative refinement} consistently improves over not only the original compression approach but also consistently outperforms a majority of baselines in terms of several metrics, offering comparable visual quality even at low bit rates. Our approach also yields better rate-distortion error using only a small number of iterative refinement steps. 
We note that our proposed decoder estimation approach is quite general and would work with any encoder. Future work will focus on integrating a neural encoder\cite{toderici2016full,balle2016end} with our SNE decoder.
\setlength{\tabcolsep}{4pt}
\begin{table}[H]
\centering\footnotesize
\caption{\footnotesize PSNR of the SNE-RNN-JP2 on the Kodak dataset (bitrate $0.3701$ bpp) as a function of $K$. 
}
\label{results:kodak_progressive}
\vspace{-0.425cm}
\begin{tabular}{|c||c|c|c|c|c|c|}
\hline
   & $K = 1$ & $K = 2$ & $K = 3$ & $K = 4$ & $K = 4$ & $K = 6$\\
  \hline
  PSNR &  28.9992&  29.2221 &  29.3007 &  29.3006 &  28.9974 &  28.8814\\
  \hline
\end{tabular}
\vspace{-0.275cm}

\caption{\footnotesize Out-of-sample results for the Kodak (bpp $0.37$), the 8-bit Compression Benchmark (CB, bpp, $0.341$), the 16-bit and 16-bit-Linear Compression Benchmark (CB) datasets (bpp $0.35$ for both), the Tecnick (bpp $0.475$), and Wikipedia (bpp $0.352$) datasets.}
\label{results:benchmarks}
\resizebox{\columnwidth}{!}{%
\begin{tabular}{|l||c|c|c|c||c|c|c|c|}
\hline
  & \multicolumn{3}{c||}{\textbf{Kodak}} & \multicolumn{3}{c|}{\textbf{CB 8-Bit}}\\
  \textbf{Model} & \textbf{PSNR} &  \textbf{SSIM} &  $MS^3IM$ & \textbf{PSNR} & \textbf{SSIM} & $MS^3IM$\\
  \hline
  \emph{JPEG} & 27.6540 & 0.7733 & 0.9291 & 27.5481 &  0.8330 & 0.9383 \\
  \emph{JPEG 2000} & 27.8370 & 0.8396 & 0.9440 & 27.7965 & 0.8362 & 0.9471 \\
  \emph{GOOG}-JPEG (\textbf{Neural}) & 27.9613 & 0.8017 & 0.9557 & 27.8458 & 0.8396 & 0.9562 \\
  
  \emph{MLP}-JPEG & 27.8325 & 0.8399 & 0.9444 & 27.8089 & 0.8371 & 0.9475 \\ 
 $\Delta$-\emph{RNN}-JPEG & 28.5093 & 0.8411 & 0.9487 & 28.0461 & 0.8403 & 0.9535 \\
 \emph{GRU}-JPEG & 28.5081 & 0.8400 & 0.9474 & 28.0446 & 0.8379 & 0.9533 \\
 \emph{LSTM}-JPEG & 28.5247 &  0.8409 & 0.9486 & 28.0461 & 0.8371 & 0.9532 \\
 \emph{LSTM}-JP2 (\textbf{Hybrid}) & {28.9321} & {0.8425} & {0.9596} & {28.0896} & {0.8389} & {0.9562} \\ 
 \emph{E2E} (\textbf{Neural}) & 28.9420  & 0.8502 & 0.9600 & 28.0999  & 0.8396 & 0.9562 \\
 \emph{SNE-RNN-SkipF}-JP2 (\textbf{Ours}) & {28.9991} & {0.8500} & {0.9602} & {28.2108} &  {0.8399} & {0.9588} \\
  \emph{SNE-RNN-SkipB}-JP2 (\textbf{Ours}) & {28.2210} &  {0.8488} & {0.9599} & {28.1001} &  {0.8377} & {0.9577} \\
  \emph{SNE-RNN-SkipBoth}-JP2 (\textbf{Ours}) & {28.8999}  & {0.8400} & {0.9599} & {28.0888} &  {0.8388} & {0.9566} \\
\emph{SNE-RNN}-JP2 (\textbf{Ours}) & \textbf{29.3008}  & \textbf{0.8508} & \textbf{0.9622} & \textbf{28.2199} & \textbf{0.8401} & \textbf{0.9600} \\
  
  \hline
    & \multicolumn{3}{c||}{\textbf{CB 16-Bit}} & \multicolumn{3}{c|}{\textbf{CB 16-Bit-Linear}}\\
  \hline
  \emph{JPEG} &27.5368 & 0.8331 & 0.9383 & 31.7522 & 0.8355 & 0.9455 \\
  \emph{JPEG 2000} & 27.7885  & 0.8391 & 0.9437 & 32.0270 & 0.8357 & 0.9471 \\
  \emph{GOOG} (\textbf{Neural}) & 27.8830 & 0.8391 & 0.9468  & 32.1275 & 0.8369 & 0.9533 \\
  
  \emph{MLP}-JPEG & 27.7762   & 0.8390 & 0.9438 & 32.0269 & 0.8356 & 0.9454 \\
  $\Delta$-\emph{RNN}-JPEG & 28.0093 & 0.8399 & 0.9471 & 32.4038 & 0.8403 & 0.9535 \\
  \emph{GRU}-JPEG & 28.0081 & 0.8392 & 0.9469 & 32.4038 & 0.8379 & 0.9533 \\
  \emph{LSTM}-JPEG & 28.0247  & 0.8310 & 0.9471 & 32.4032  & 0.8371 & 0.9532 \\
  \emph{LSTM}-JP2 (\textbf{Hybrid}) & {28.1307}  & {0.8425} & {0.9496} & {32.4998} & {0.8382} & {0.9541} \\
   \emph{E2E} (\textbf{Neural}) & 28.2440 & 0.8426 & 0.9498  & 32.5010 & 0.8387 & 0.9540 \\
  \emph{SNE-RNN-SkipF}-JP2 (\textbf{Ours}) & {28.2441}  & {0.8426} & {0.9499} & {32.5000} &  {0.8381} & {0.9541} \\
  \emph{SNE-RNN-SkipB}-JP2 (\textbf{Ours}) & {28.1007}  & {0.8422} & {0.9492} & {32.4992} &  {0.8382} & {0.9542} \\
  \emph{SNE-RNN-SkipBoth}-JP2 (\textbf{Ours}) & {28.1006} & {0.8412} & {0.9491} & {32.4981} & {0.8381} & {0.9538} \\
\emph{SNE-RNN}-JP2 (\textbf{Ours}) & \textbf{29.4471} &  \textbf{0.8430} & \textbf{0.9510} & \textbf{32.6019} & \textbf{0.8399} & \textbf{0.9559} \\
  
  \hline
  & \multicolumn{3}{c||}{\textbf{Tecnick}} & \multicolumn{3}{c|}{\textbf{Wikipedia}}\\
  \hline
  \emph{JPEG} &  30.7377   & 0.8682 &  0.9521 & 28.7724 & 0.8290 &  0.9435\\
  \emph{JPEG 2000} & 31.2319  & 0.8747 & 0.9569 & 29.1545  & 0.8382 & 0.9495 \\
  \emph{GOOG} (\textbf{Neural}) & 31.5030 & 0.8814 & 0.9608 & 29.2209 &  0.8406 &  0.9520 \\
   
  \emph{MLP}-JPEG & 31.2287 & 0.8746 &  0.9571 & 29.1547  &  0.8383 &  0.9497 \\
  $\Delta$-\emph{RNN}-JPEG &  31.5411  &  0.8821 &  0.9609 & 29.2772 & 0.8403  & 0.9519\\
  \emph{LSTM}-JPEG & 31.5616 & 0.8820 & 0.9609 & 29.2771 & 0.8403 & 0.9519\\
  \emph{LSTM}-JP2 (\textbf{Hybrid}) & {31.6962} &  {0.8834} & {0.9619} & {29.3228}  & {0.8411} & {0.9526} \\
  \emph{E2E} (\textbf{Neural}) & 31.7000  & 0.8836 & 0.9620 & 29.3227  &  0.8412 &  0.9526 \\
  \emph{SNE-RNN-SkipF}-JP2 (\textbf{Ours}) & {31.6999}  & {0.8835} & {0.9621} &{29.3226} &  {0.8413} & {0.9525} \\
  \emph{SNE-RNN-SkipB}-JP2 (\textbf{Ours})& {31.6950}  & {0.8830} & {0.9600} & {29.3221} &  {0.8402} & {0.9524} \\
  \emph{SNE-RNN-SkipBoth}-JP2 (\textbf{Ours}) & {31.6944}& {0.8822} & {0.9589} & {29.3001} &  {0.8345} & {0.9501} \\
\emph{SNE-RNN}-JP2 (\textbf{Ours}) & \textbf{32.7124} &  \textbf{0.8841} & \textbf{0.9622} & \textbf{29.9334} &  \textbf{0.8413} & \textbf{0.9528} \\
  \hline
\end{tabular}
}
\end{table}
\setlength{\tabcolsep}{1.4pt}


\Section{References}
\bibliographystyle{IEEEbib}
\bibliography{neural_refs}

\end{document}